# НЕОЛОГИЗМЫ В СОЦИАЛЬНОЙ СЕТИ ФЕЙСБУК

**Муравьев Н. А.** (nikita.muraviev@gmail.com)

ООО «Лаборатория Цифрового Общества», Москва, Россия;
МГУ, Отделение Теоретической и Прикладной
Лингвистики, Москва, Россия

**Панченко А. И.** (a.panchenko@digsolab.com)

ООО «Лаборатория Цифрового Общества», Москва, Россия;
Лувенский католический Университет, Лувен, Бельгия

**Объедков С. А.** (sergei.obj@digsolab.com)

ООО «Лаборатория Цифрового Общества», Москва, Россия;
НИУ ВШЭ, Отделение прикладной математики
и информатики, Москва, Россия

Исследование проведено на материале наиболее частотных словоформ социальной сети Фейсбук, отсутствующих в словарях. Главная задача исследования состояла в анализе новообразований русского языка с точки зрения наиболее характерных словообразовательных моделей и частей речи, а также адаптации заимствований. Результат нашей работы — словарь из 168 неологизмов и его лингвистический анализ по типу заимствования.

**Ключевые слова:** неологизм, заимствование, несловарное слово, словообразование, социальная сеть

# NEOLOGISMS ON FACEBOOK

**Muravyev N. A.** (nikita.muraviev@gmail.com)

Digital Society Laboratory LLC, Moscow, Russia;
Moscow State University, Faculty of Theoretical and Applied Linguistics, Moscow, Russia

**Panchenko A. I.** (a.panchenko@digsolab.com)

Digital Society Laboratory LLC, Moscow, Russia;
Universite catholique de Louvain, Louvain-la-Neuve, Belgium

**Obiedkov S. A.** (sergei.obj@digsolab.com)

Digital Society Laboratory LLC, Moscow, Russia;
National Research University Higher School of Economics, Department of Applied Mathematics and Information Science, Moscow, Russia

In this paper, we present a study of neologisms and loan words frequently occurring in Facebook user posts. We have collected a dataset of over 573 million posts written during 2006–2013 by Russian-speaking Facebook users. From these, we have built a vocabulary of most frequent lemmatized words missing from the Opencorpora dictionary (http://opencorpora.org/dict.php) the assumption being that many such words have entered common use only recently. This assumption is certainly not true for all the words extracted in this way; for that reason, we manually filtered the automatically obtained list in order to exclude non-Russian or incorrectly lemmatized words, as well as words recorded by other dictionaries or those occurring in pre-2000 texts from the Russian National Corpus (http://www.ruscorpora.ru). The result is a list of 168 words that can potentially be considered neologisms.

We present an attempt at an etymological classification of these neologisms (unsurprisingly, most of them have recently been borrowed from English, but there are also quite a few new words composed of previously borrowed stems) and identify various derivational patterns. We also classify words into several large thematic areas, "internet", "marketing", and "multimedia" being among those with the largest number of words.

We consider our results preliminary, but believe that, together with the word base collected in the process, they can serve as a starting point in further studies of neologisms and lexical processes that lead to their acceptance into the mainstream language.

**Keywords:** neologism, loan word, non-vocabulary, derivation, social network

Muravyev N. A., Panchenko A. I., Obiedkov S. A.

## 1. Введение

Систематическое изучение неологизмов и заимствованных слов как отдельной разновидности неологизмов должно дать ответ на различные вопросы, связанные с тем, каким образом меняется лексический состав языка с течением времени и, в частности, по каким моделям и с помощью каких средств интегрируются иноязычные и новообразованные лексические единицы в язык, и как происходит их адаптация. Под моделями мы понимаем как морфологические паттерны словообразования и словоизменения, так его частеречную принадлежность и способность употребляться в различных позициях в предложении.

Помимо формальной стороны вопроса, интерес также представляет семантика лексем, а именно развитие наиболее актуальных семантических полей, лексических отношений внутри них и полисемии лексем. Кроме того, не следует обходить вниманием и такие моменты как, например, вариативность в орфографии заимствованных слов.

Что касается других постоянно пополняющихся классов слов, таких как, например, диалектная лексика, жаргонизмы или имена собственные, они остаются за рамками нашего исследования. Таким образом, объектом нашего исследования являются новообразованные слова, которые необходимо кодифицировать.

Модели словообразования и заимствования представляют не только теоретический, но и прикладной интерес. Так, большинство современных морфологических анализаторов достаточно плохо справляются с обработкой несловарных слов. Знание о том, как образуются новые слова, могло бы помочь в создании более точных систем автоматической обработки текстов. Основной задачей нашей работы было построение первичной словарной базы наиболее частотных неологизмов, используемых в социальных сетях, которая бы стала отправной точкой для подобного рода исследований.

## 2. История вопроса

Заимствованиям посвящен широкий спектр теоретических работ и типологических работ. В числе фундаментальных трудов по теории заимствованных слов можно назвать классическую работу [Haugen 1950], а также различные исследования последних лет [Capuz 1997, Winter-Froemel 2008, Peperkamp&Dupoux 2003, LaCharité&Paradis 2005]. Существует также немало конкретно-языковых исследований, например, [Hall & Hamann 2003, Volland 1986] по заимствованиям в немецком языке или [Heinemann 2003] — в итальянском языке. Что касается русского языка, то прежде всего стоит упомянуть фундаментальную работу Л. П. Крысина «Иноязычные слова в современном русском языке», а также работы [Брейтер 1997, Дьяков 2003], посвященные англицизмам, и монографию [Маринова 2013]. Неологизмам в узком смысле, то есть словам, образованным от уже имеющихся корней, уделяется несколько меньше внимания. Среди работ по данной тематике можно отметить исследования на материале английского [Fisher 1998], французского [Cougnon 2010] и каталанского [Estopa 2011] языков. Также можно отметить исследование [Ahmad 1991], посвящённое лексическим изменениям в современном английском языке, где рассматриваются как заимствования, так и неологизмы.

## 3. Анализ несловарных слов социальной сети Фейсбук

### 3.1. Корпус постов и комментариев русскоязычного Фейсбука

Словарь неологизмов, описанный в данной статье, был построен полуавтоматически из корпуса анонимизированных постов и комментариев Фейсбука[1]. Корпус был получен Лабораторией Цифрового Общества[2] из русскоязычного публичного сегмента социальной сети при помощи программного интерфейса (API) Фейсбука[3]. В набор данных попали сообщения пользователей с «открытым» профилем. Такие профили доступны для чтения для всех других пользователей социальной сети. Построение подобных выборок данных в исследовательских целях распространено в научном сообществе [Catanese et al 2012].

Извлечение неологизмов было произведено из 573 миллионов анонимизированных постов и сообщений 3,2 млн. пользователей социальной сети. В данном эксперименте рассматривались только тексты на русском языке. Язык каждого из входных текстов был определён автоматически при помощи модуля *langid.py* [Lui & Baldwin 2012]. В Таблице 1 приведены основные параметры корпуса текстов русскоязычного Фейсбука, использованного в данной работе. Исследуемый корпус содержит *посты* пользователей за период с 2006 по 2013 год. Первый пост в корпусе датируется 5 августа 2006 года, в то время как последний пост был написан 13 ноября 2013 года. Однако подавляющее большинство постов написано в период с 2011 по 2013 год.

Согласно официальному сайту данной социальной сети [18], всемирная аудитория Фейсбука насчитывала около 1,19 миллиарда активных пользователей в сентябре 2013 года. К сожалению, распределение по странам не представлено на официальной странице. Согласно сайту Internet World Stats [19], количество пользователей Фейсбука в России на конец 2012 года приблизилось к 8 миллионам человек. Таким образом, рассматриваемый набор данных представляет около 40% пользователей русскоязычного Фейсбука 2012 года.

---

[1] http://www.facebook.com

[2] http://www.digsolab.ru

[3] https://developers.facebook.com/tools/explorer

**Таблица 1.** Статистика корпуса русскоязычных постов и комментариев Фейсбука

| Параметр | Значение |
|---|---|
| **Количество анонимизированных пользователей** | **3 190 813** |
| Язык | Русский |
| Количество постов | 426 089 762 |
| Количество комментариев | 147 140 265 |
| **Количество текстов (посты и комментарии)** | **573 230 027** |
| Количество словоформ в постах | 20 775 837 467 |
| Количество словоформ в комментариях | 2 759 777 659 |
| **Количество словоформ (посты и комментарии)** | **23 535 615 126** |
| Средняя длина поста, словоформ | 49 |
| Средняя длина комментария, словоформ | 19 |

### 3.2. Построение словаря неологизмов

Мы использовали следующий полуавтоматический подход для построения словаря неологизмов из корпуса текстов, описанного выше:

**1. Построение частотного словаря.** Каждый текст (пост или комментарий) был токенизирован и лемматизирован при помощи морфологического анализатора, основанного на словаре *АОТ*[4], который, в свою очередь, основан на словаре Зализняка [Зализняк 1977]. Мы использовали собственную *MapReduce* реализацию модуля построения частотного словаря, основанную на модуле морфологического анализа *RussianMorphology*[5].

**2. Морфологическая разметка частотного словаря.** Полученный частотный словарь был аннотирован при помощи морфологического анализатора *PyMorphy*[6]. Каждой лемме была присвоена часть речи. Кроме этого, для каждой леммы было указано, входит ли она в словарь *OpenCorpora*[7]. Данный морфологический словарь основан на словаре *АОТ* и включает в себя информацию о 388 790 леммах и 5 094 925 словоформах. Построенный словарь несловарных слов Фейсбука был отфильтрован по частоте. К сожалению, на этом этапе произошел отсев новых слов, омонимичных существующим словам русского языка. Примером такого слова, оставшегося за пределами исследования, является используемое в данной статье слово «пост», относительно недавно заимствованное из английского языка в значении «заметка, размещенная пользователем в социальной сети».

---

[4] http://www.aot.ru/

[5] https://code.google.com/p/russianmorphology/

[6] https://bitbucket.org/kmike/pymorphy2

[7] http://opencorpora.org/dict.php

3. **Фильтрация словаря экспертами.** Полученный частотный словарь несловарных слов оказался крайне «зашумленным». К примеру, в него вошли нерусские слова, неверно лемматизированные слова и другие артефакты автоматической процедуры описанной выше. Следующие несловарные слова оказались среди наиболее частотных в нашем словаре: ть, нибыть, гый, санкт, що, ул, пр, нью, грн, ца, рожение, т.д, від, україні, вебинара, дтпа, кя, свый, плэйкастый, сегода, др, бй, квна, т.е, кг, млма, гр, бо, який, ра, ка, т.к, бть, чи, ск, холти. Для исправления ситуации нами была проведена ручная фильтрация 10 000 наиболее частотных слов. В результате данного этапа был получен словарь, из которого мы отобрали 624 наиболее частотных несловарных слова[8].

4. **Лингвистическая фильтрация словаря.** В результате фильтрации на предыдущем этапе был получен гораздо менее «зашумленный» список несловарных слов, которые часто употребляются пользователями социальной сети. Однако в данный список попало большое количество несловарных слов, которые нельзя отнести к неологизмам. Во-первых, в список попало много имен собственных: географические объекты, имена, фамилии, названия организаций и т.п. Во-вторых, в словарь попала распространенная сниженная лексика. Наконец, многие слова, такие как «авиаперелет», «переориентироваться» и «однодневный», не содержатся в словаре *OpenCorpora*, однако присутствуют в других словарях и не являются неологизмами. Поэтому была произведена дополнительная ручная фильтрация списка несловарных слов авторами статьи, в результате которой были удалены слова, обнаруженные в поисковой системе *Яндекс.Словари*[9] и в Национальном Корпусе Русского Языка (НКРЯ) [Плунгян 2003]. Система *Яндекс.Словари* производит поиск по 10 словарям русского языка, 50 энциклопедиям и 22 словарям иностранных языков. НКРЯ на февраль 2013 года содержал 335 076 текста (364 881 378 словоформ), которые можно разделить на две группы: современные письменные тексты середины XX — начала XXI века и ранние тексты середины XVIII — середины XX века. Мы использовали поиск по всем текстам НКРЯ до 2000 года. Результат данного этапа — список из 168 популярных неологизмов, извлеченных из корпуса текстов социальной сети.

5. **Лингвистический анализ словаря неологизмов.** Наконец, мы классифицировали полученный список неологизмов по типу заимствования, типу словообразования и модели словообразования (см. таблицу 2, приложение и следующие разделы статьи).

---

[8] Список несловарных слов и неологизмов: https://www.dropbox.com/s/miwknzucui13l60/neo-facebook.xlsx

[9] http://slovari.yandex.ru

**Таблица 2.** 168 неологизмов Фейсбука, упорядоченных по типу заимствования

| Неологизм | Тип заим­ствования | Тип словообра­зования | Модель словообра­зования |
|---|---|---|---|
| сексодром | Англицизм | | |
| айпад, айфон, алерт, байк, бейдж, билдер, блоггинг, брейн, брендинг, вау, виджет, девелопер, демотиватор, дресс, инфо, ка­вер, караванер, клаб, корпоратив, комент, коммент, коучинг, лайт, лайф, мем, ноут, паблик, перфоманс, плиз, праймериз, принт, продакшн, промо, райдер, ребрен­динг, рекрутинг, репост, ретвит, реферал, ритейл, ритейлер, роутер, сиквел, скайп, скрин, сорри, стайл, стор, твитер, твиттер, тизер, трекер, треш, трэш, фейк, форсайт, фреш, фэшн, хайп, холдем, чарт, шутер | Англицизм | Исх 1 корень | |
| битрейт, бумбокс, геймплей, дабстеп, дедлайн, инфомаркетинг, клипарт, ко­пирайтинг, никнейм, оффлайн, плагин, плейлист, плэйкаст, подкаст, рингтон, стартапер, топфейс, фейсбук, флешмоб, флэшмоб, фолловер, форекс, фрилансер, фэйсбук, хардкор, ютуб, ютюб | Англицизм | Исх 2 корня | |
| декупаж | Галлицизм | | |
| жжот, капец, мульт, мда, медвед, пипец, ппц, секстиль | Исконное | | |
| госуслуга | Исконное | Композит | ST-ST |
| единорос | Исконное | Композит | ST-о-ST |
| всечь | Исконное | Префикс | в-ST |
| нафиг, нахер, нахрен | Исконное | Префикс | на-ST |
| предзаказ | Исконное | Префикс | пред-ST |
| заценить | Исконное | Префикс+суффикс | за-ST-и |
| офигевать | Исконное | Префикс+суффикс | о-ST-ева |
| прокремлевский | Исконное | Префикс+суффикс | про-ST-ск |
| бухарь | Исконное | Суффикс | ST-арь |
| улыбизм | Исконное | Суффикс | ST-изм |
| приколист | Исконное | Суффикс | ST-ист |
| личка, печенька, ржака | Исконное | Суффикс | ST-к |
| ржачный, улетный | Исконное | Суффикс | ST-н |
| херня | Исконное | Суффикс | ST-нь |
| пристройство | Исконное | Суффикс | ST-ств |
| ржач | Исконное | Суффикс | ST-ч |
| адчайший | Исконное | Суффикс | ST-ч-айш |
| днюха | Исконное | Суффикс | ST-юх |
| вкусняшка | Исконное | Суффикс | ST-яшк |
| евроинтеграция, инфографика, инфопро­дукт, телепроект, фотопроект, фотосту­дия, видеорепортаж | Из заимств. корней | Композит | ST-ST |

| Неологизм | Тип заим-ствования | Тип словообра-зования | Модель словообра-зования |
|---|---|---|---|
| аудиокнига, вконтакт, мультиварка, нар-деп, фотолента, фотоотчет, фотопамять, фотоподборка, фотоприкол, фотошкола | Смешанное | Композит | ST-ST |
| лохотрон, файлообменник | Смешанное | Композит | ST-о-ST |
| перепост | Смешанное | Префикс | пере-ST |
| предстарт | Смешанное | Префикс | пред-ST |
| забанить, запостить | Смешанное | Префикс+суффикс | за-ST-и |
| зацикливаться | Смешанное | Префикс+суффикс | за-ST-ива |
| перепостить | Смешанное | Префикс+суффикс | пере-ST-и |
| лайкать | Смешанное | Суффикс | ST-а |
| культурить, постить, твитить | Смешанное | Суффикс | ST-и |
| анимировать | Смешанное | Суффикс | ST-ирова |
| аватарка, гифка, флешка | Смешанное | Суффикс | ST-к |
| реферальный | Смешанное | Суффикс | ST-н |
| планшетник, цитатник | Смешанное | Суффикс | ST-ник |
| имхонуть, лайкнуть | Смешанное | Суффикс | ST-ну |
| брендовый, драйвовый | Смешанное | Суффикс | ST-ов |
| форумок | Смешанное | Суффикс | ST-ок |
| суперский | Смешанное | Суффикс | ST-ск |
| позитивчик | Смешанное | Суффикс | ST-чик |
| креативщик | Смешанное | Суффикс | ST-щик |

### 3.3. Классификация наиболее частотных несловарных слов

Для понимания языковых изменений важно знать, откуда произошло слово, каким способом оно образовано, каковы его морфологические и синтаксические свойства, и как оно используется. Руководствуясь данным соображением, мы классифицировали полученный список слов по пяти основаниям: *тип заимствования*, *часть речи*, *тип словообразования*, *словообразовательная модель* и *тематика* (см. таблицу в Приложении). Именно эти характеристики мы считаем наиболее показательными в анализе современных заимствований и неологизмов.

По типу заимствования слова из списка разделены на пять классов: исконные, англицизмы, галлицизмы, слова с иноязычными корнями и смешанные слова. Исконными словами считаются те лексемы, которые образованы от общеславянских корней, к англицизмам и галлицизмам причислены новейшие заимствования из английского и французского языков, к словам из заимствованных корней отнесены композиты, составленные из заимствований более раннего времени, тогда как к смешанным словам отнесены слова, состоящие из русскоязычных и иноязычных морфем.

Части речи представлены шестью основными классами: существительное (N), существительное-модификатор (Nmod, подробнее см. раздел 4), прилагательное (Adj), глагол (V), наречие (Adv) и междометие (Interj).

Что касается словообразовательных характеристик, то классификация по типу словообразования соответствует традиционно-грамматическому делению лексем на образованные с помощью суффикса, префикса, префикса и суффикса и словосложения. Также при непроизводных заимствованных словах дополнительно указывается количество корней в языке-источнике (один или два). Словообразовательные модели представляют собой записанную через дефис комбинацию основы слова (ST) и участвующих в словообразовании элементов: префиксов, суффиксов и др.

Наконец, тематика слов как основание для классификации делит слова данного списка на семантические поля, к которым они относятся. Наиболее многочисленными по составу являются семантические поля «интернет» (*офлайн, браузерный*), «оценка» (*суперский, треш/трэш*), «маркетинг» (*реферал, продакшн*) и «мультимедиа» (*фотопроект, плэйкаст*). Следует оговориться, что данное деление на семантические поля условно, поскольку одно и то же слово может принадлежать к разным полям. Поэтому в спорных случаях мы как правило отдавали предпочтение более крупному полю.

## 4. Полученные результаты и некоторые наблюдения

Предсказуемо большее число неологизмов составляют слова, полностью заимствованные из других языков (93 слова), однако любопытно то, что практически все они, кроме галлицизма *декупаж*, являются словами английского происхождения. Кроме того, есть также семь слов, содержащих иноязычные корни (всего 43), из которых восемь слов составлены из иностранных корней и еще 35 имеют смешанное происхождение, т. е. содержат как иноязычные, так и русские морфемы. Остальные слова (всего 32) являются исконно русскими или калькированными с других языков при помощи русских корней. Наиболее активно пополняемым семантическим полем, по нашим данным, является поле «интернет», представленное практически исключительно англицизмами и образованными от них словами (35 слов). Второе место занимает поле «оценка» (25 слов), поскольку данное семантическое поле является, по всей очевидности, одним из наиболее востребованных, в особенности с появлением социальных сетей и возможностью комментировать выкладываемый в них материал. Стоит также отметить, что к данному полю относится ровно половина исконно русских неологизмов из нашего списка (16 слов). Другие два активно развивающихся поля — «маркетинг» (18 слов) и «техника» (14 слов) — также представлены преимущественно английскими заимствованиями. Наконец, основу еще одного обширно представленного в материале поля «мультимедиа» (15 слов) составляют смешанные слова или композиты из иноязычных корней.

По своей частеречной принадлежности абсолютное большинство слов (123) являются существительными, а также имеется 15 глаголов, 8 прилагательных, 4 междометия и 3 наречно-предикативных слова. Кроме того, имеется особый класс существительных-модификаторов (всего 15 слов), таких как, например, *брейн, лайф* и *фэшн*. Такие единицы в основном англоязычного происхождения в силу незначительной степени адаптированности употребляются почти всегда в сочетаниях с другими существительными, модифицируя их по принципу английских конструкций типа «stone wall»: *брейн-система, лайф-коуч, фэшн-индустрия* (нередко встречается раздельное написание). Однако среди этих слов встречаются и такие, которые в некоторых случаях могут употребляться самостоятельно, например *байк* и *трэш/треш* (ср. *байк-центр* и *трэш-комедия*). Такие случаи объединены в отдельный промежуточный класс (Nmod/N). В целом такая модель является новой для русского языка.

Что же касается морфологии данных слов, то всего в списке имеется 101 непроизводное и 67 производных слов. При этом стоит отметить, что среди непроизводных слов только восемь являются исконными, тогда как среди производных их число равняется 24. Наиболее распространенным способом словообразования является суффиксация (33). К числу продуктивных суффиксальных моделей можно отнести субстантивную модель с суффиксом -к- (6 слова), в равной степени образующую слова от русских и иноязычных корней, а также глагольную модель с суффиксом -и- (3 слова), имеющую префиксально-суффиксальный коррелят (4 слова). Отдельного внимания заслуживает субстантивная модель с суффиксом -ч-, поскольку является, по всей очевидности, инновацией. К данной модели из списка относится слово *ржач*, однако нам известны и другие случаи, например слова *срач* и *махач*. Других новых моделей на данном материале не зафиксировано. Вторым по распространенности способом является словосложение, представленное, за исключением слов *единорос* и *госуслуга*, словами иноязычного и смешанного происхождения. Из оставшихся префиксального и префиксально-суффиксального способа словообразования (по 7 слов) наиболее продуктивными являются, соответственно, модель с префиксом на- и с префиксом за- и суффиксом -и- (по 3 слова).

Если обобщить изложенные выше данные, то в настоящее время наблюдается активный поток заимствований из английского языка, окончательно закрепившего за собой статус языка международного общения, о чем свидетельствует абсолютное превосходство англицизмов над всеми остальными неологизмами из нашего списка. Практически все остальные слова возникают в результате комбинаций уже существующих в языке исконных и иноязычных морфем. Основными активно обновляющимися в лексическом плане сферами являются мультимедиа- и интернет-технологии, а также торговля, что свидетельствует о существенной роли данных областей в современной жизни. Что характерно, обновление данных полей, кроме поля «мультимедиа», происходит практически исключительно за счет англицизмов. Неологизмы с исконными корнями же возникают прежде всего в оценочной сфере в силу растущей потребности выражать собственное мнение о материалах, выкладываемых в социальные сети. Большинство как исконных, так и заимствованных неологизмов являются существительными, поскольку данная часть речи является наиболее подходящей для описания новых для языка реалий. Заимствованные неологизмы, являясь по большей части существительными, преимущественно непроизводные, и только часть слов, адаптированных в системе русского языка, содержит словообразовательные морфемы. Исконные слова же, закономерным образом, почти все являются производными, поскольку единственным, помимо заимствования способом образования неологизмов является деривация.

## 5. Заключение

В результате нашей работы полуавтоматическим способом был получен словарь из 168 неологизмов русскоязычного сегмента социальной сети Фейсбук. Найденные неологизмы представляют собой наиболее частотную часть списка и активно используются и в других социальных сетях и платформах в интернете. Таким образом, мы создали словарную базу, которая может быть использована для различных исследований, касающихся как возникновения неологизмов, так и процесса заимствования и адаптации заимствованных слов в языке.

Кроме того, мы предложили классификацию полученного материала по типу заимствования, тематике, части речи, типу словообразования и модели словообразования. Данная классификация позволяет получить первоначальное представление о специфике морфологии и синтаксиса новейших лексических единиц языка и происходящих в наше время лексических процессах, дает возможность обратить внимание на актуальные сферы использования современной лексики и сравнить свойства исконных и заимствованных слов.

Наконец, мы надеемся, что результаты наших исследований помогут улучшить понимание принципов автоматической обработки и анализа слов, отсутствующих в словарях, что должно улучшить состояние технологий обработки текста.

## Литература

## Приложение 1. Словарь неологизмов русскоязычного Фейсбука

| # | Слово | Часть речи | Тематика | Тип заимствования | Тип словообразования | Модель словообразования | Частота |
|---|---|---|---|---|---|---|---|
| 1 | лайкать | V | интернет | Смешанное | Суффикс | ST-а | 34222 |
| 2 | вконтакт | N | интернет | Смешанное | Композит | ST-ST | 1134119 |
| 3 | перепост | N | интернет | Смешанное | Префикс | пере-ST | 426927 |
| 4 | постить | V | интернет | Смешанное | Суффикс | ST-и | 61998 |
| 5 | твитить | V | интернет | Смешанное | Суффикс | ST-и | 100988 |
| 6 | файлообменник | N | интернет | Смешанное | Композит | ST-о-ST | 27638 |
| 7 | аватарка | N | интернет | Смешанное | Суффикс | ST-к | 22432 |
| 8 | забанить | V | интернет | Смешанное | Префикс+суффикс | за-ST-и | 18400 |

| # | Слово | Часть речи | Тематика | Тип заимствования | Тип словообразования | Модель словообразования | Частота |
|---|---|---|---|---|---|---|---|
| 9 | запостить | V | интернет | Смешанное | Префикс+суффикс | за-ST-и | 21971 |
| 10 | перепостить | V | интернет | Смешанное | Префикс+суффикс | пере-ST-и | 212364 |
| 11 | цитатник | N | интернет | Смешанное | Суффикс | ST-ник | 59566 |
| 12 | лайкнуть | V | интернет | Смешанное | Суффикс | ST-ну | 68457 |
| 13 | имхонуть | V | интернет | Смешанное | Суффикс | ST-ну | 17395 |
| 14 | форумок | N | интернет | Смешанное | Суффикс | ST-ок | 26937 |
| 15 | личка | N | интернет | Исконное | Суффикс | ST-к | 247497 |
| 16 | блоггинг | N | интернет | Англицизм | | | 22332 |
| 17 | комент | N | интернет | Англицизм | | | 30312 |
| 18 | коммент | N | интернет | Англицизм | | | 117196 |
| 19 | никнейм | N | интернет | Англицизм | | | 17702 |
| 20 | оффлайн | N | интернет | Англицизм | | | 21614 |
| 21 | паблик | N | интернет | Англицизм | | | 48587 |
| 22 | репост | N | интернет | Англицизм | | | 144676 |
| 23 | ретвит | N | интернет | Англицизм | | | 29336 |
| 24 | скайп | N | интернет | Англицизм | | | 426877 |
| 25 | твитер | N | интернет | Англицизм | | | 22751 |
| 26 | твиттер | N | интернет | Англицизм | | | 578503 |
| 27 | топфейс | N | интернет | Англицизм | | | 161399 |
| 28 | трекер | N | интернет | Англицизм | | | 45888 |
| 29 | фейсбук | N | интернет | Англицизм | | | 377884 |
| 30 | фолловер | N | интернет | Англицизм | | | 22124 |
| 31 | фэйсбук | N | интернет | Англицизм | | | 62267 |
| 32 | ютуб | N | интернет | Англицизм | | | 83074 |
| 33 | ютюб | N | интернет | Англицизм | | | 24043 |
| 34 | битрейт | N | интернет | Англицизм | | | 58676 |
| 35 | инфографика | N | интернет | Англицизм | | | 91729 |
| 36 | культурить | V | культура | Смешанное | Суффикс | ST-и | 24073 |
| 37 | секстиль | N | культура | Исконное | | | 117415 |
| 38 | дабстеп | N | культура | Англицизм | | | 18255 |
| 39 | демотиватор | N | культура | Англицизм | | | 22432 |
| 40 | кавер | N | культура | Англицизм | | | 72386 |
| 41 | клипарт | N | культура | Англицизм | | | 23663 |
| 42 | мем | N | культура | Англицизм | | | 38912 |
| 43 | перфоманс | N | культура | Англицизм | | | 23905 |
| 44 | плэйкаст | N | культура | Англицизм | | | 367575 |
| 45 | сиквел | N | культура | Англицизм | | | 17685 |
| 46 | флешмоб | N | культура | Англицизм | | | 84966 |
| 47 | флэшмоб | N | культура | Англицизм | | | 34780 |
| 48 | предстарт | N | маркетинг | Смешанное | Префикс | пред-ST | 23441 |
| 49 | реферальный | Adj | маркетинг | Смешанное | Суффикс | ST-н | 93330 |
| 50 | предзаказ | N | маркетинг | Исконное | Префикс | пред-ST | 22915 |
| 51 | инфопродукт | N | маркетинг | Из заимств. корней | Композит | ST-ST | 17824 |
| 52 | алерт | N | маркетинг | Англицизм | | | 19949 |
| 53 | брендинг | N | маркетинг | Англицизм | | | 33457 |
| 54 | инфомаркетинг | N | маркетинг | Англицизм | | | 17928 |
| 55 | продакшн | N | маркетинг | Англицизм | | | 33134 |
| 56 | промо | Nmod | маркетинг | Англицизм | | | 134124 |
| 57 | ребрендинг | N | маркетинг | Англицизм | | | 19333 |
| 58 | реферал | N | маркетинг | Англицизм | | | 103650 |
| 59 | ритейл | N | маркетинг | Англицизм | | | 21971 |
| 60 | ритейлер | N | маркетинг | Англицизм | | | 23151 |
| 61 | стартапер | N | маркетинг | Англицизм | | | 21100 |
| 62 | стор | Nmod | маркетинг | Англицизм | | | 36337 |
| 63 | тизер | N | маркетинг | Англицизм | | | 49417 |
| 64 | форекс | N | маркетинг | Англицизм | | | 186372 |
| 65 | хайп | N | маркетинг | Англицизм | | | 31586 |
| 66 | анимировать | V | мультимедиа | Смешанное | Суффикс | ST-ирова | 33222 |

| # | Слово | Часть речи | Тематика | Тип заимствования | Тип словообразования | Модель словообразования | Частота |
|---|---|---|---|---|---|---|---|
| 67 | гифка | N | мультимедиа | Смешанное | Суффикс | ST-к | 28293 |
| 68 | фотоотчет | N | мультимедиа | Смешанное | Композит | ST-ST | 129303 |
| 69 | фотопамять | N | мультимедиа | Смешанное | Композит | ST-ST | 67254 |
| 70 | фотоподборка | N | мультимедиа | Смешанное | Композит | ST-ST | 25999 |
| 71 | фотоприкол | N | мультимедиа | Смешанное | Композит | ST-ST | 85613 |
| 72 | мульт | N | мультимедиа | Исконное | | | 76768 |
| 73 | видеорепортаж | N | мультимедиа | Из заимств. корней | Композит | ST-ST | 17665 |
| 74 | телепроект | N | мультимедиа | Из заимств. корней | Композит | ST-ST | 26985 |
| 75 | фотолента | N | мультимедиа | Из заимств. корней | Композит | ST-ST | 18920 |
| 76 | фотопроект | N | мультимедиа | Из заимств. корней | Композит | ST-ST | 45528 |
| 77 | фотостудия | N | мультимедиа | Из заимств. корней | Композит | ST-ST | 31041 |
| 78 | фотошкола | N | мультимедиа | Из заимств. корней | Композит | ST-ST | 17942 |
| 79 | лайф | Nmod | мультимедиа | Англицизм | | | 28368 |
| 80 | скрин | N | мультимедиа | Англицизм | | | 23283 |
| 81 | брендовый | Adj | одежда | Смешанное | Суффикс | ST-ов | 26893 |
| 82 | бейдж | N | одежда | Англицизм | | | 17981 |
| 83 | дресс | Nmod | одежда | Англицизм | | | 32026 |
| 84 | принт | N | одежда | Англицизм | | | 81120 |
| 85 | фэшн | Nmod | одежда | Англицизм | | | 32895 |
| 86 | лохотрон | N | оценка | Смешанное | Композит | ST-о-ST | 30760 |
| 87 | драйвовый | Adj | оценка | Смешанное | Суффикс | ST-ов | 24135 |
| 88 | суперский | Adj | оценка | Смешанное | Суффикс | ST-ск | 19754 |
| 89 | позитивчик | N | оценка | Смешанное | Суффикс | ST-чик | 22556 |
| 90 | нафиг | Adv/Pred | оценка | Исконное | Префикс | на-ST | 39668 |
| 91 | приколист | N | оценка | Исконное | Суффикс | ST-ист | 69330 |
| 92 | нахер | Adv/Pred | оценка | Исконное | Префикс | на-ST | 17553 |
| 93 | нахрен | Adv/Pred | оценка | Исконное | Префикс | на-ST | 22916 |
| 94 | ржака | N | оценка | Исконное | Суффикс | ST-к | 43288 |
| 95 | ржачный | Adj | оценка | Исконное | Суффикс | ST-н | 22716 |
| 96 | жжот | V | оценка | Исконное | | | 31450 |
| 97 | капец | N | оценка | Исконное | | | 27650 |
| 98 | зацепить | V | оценка | Исконное | Префикс+суффикс | за-ST-и | 200777 |
| 99 | пипец | N | оценка | Исконное | | | 51314 |
| 100 | улетный | Adj | оценка | Исконное | Суффикс | ST-н | 19758 |
| 101 | ппц | N | оценка | Исконное | | | 29883 |
| 102 | херня | N | оценка | Исконное | Суффикс | ST-нь | 28836 |
| 103 | офигевать | V | оценка | Исконное | Префикс+суффикс | о-ST-ева | 26990 |
| 104 | ржач | N | оценка | Исконное | Суффикс | ST-ч | 32035 |
| 105 | адчайший | Adj | оценка | Исконное | Суффикс | ST-ч-айш | 22432 |
| 106 | премиум | Nmod | оценка | Англицизм | | | 58276 |
| 107 | треш | Nmod/N | оценка | Англицизм | | | 17421 |
| 108 | трэш | Nmod/N | оценка | Англицизм | | | 27304 |
| 109 | фейк | N | оценка | Англицизм | | | 24691 |
| 110 | хардкор | N | оценка | Англицизм | | | 20181 |
| 111 | вкусняшка | N | питание | Исконное | Суффикс | ST-яшк | 30777 |
| 112 | фреш | Nmod/N | питание | Англицизм | | | 18830 |
| 113 | нардеп | N | политика | Смешанное | Композит | ST-ST | 44061 |
| 114 | госуслуга | N | политика | Исконное | Композит | ST-ST | 17942 |
| 115 | единорос | N | политика | Исконное | Композит | ST-о-ST | 17405 |
| 116 | прокремлевский | Adj | политика | Исконное | Префикс+суффикс | про-ST-ск | 18001 |
| 117 | евроинтеграция | N | политика | Из заимств. корней | Композит | ST-ST | 16984 |
| 118 | праймериз | N | политика | Англицизм | | | 23514 |
| 119 | зацикливаться | V | психология | Смешанное | Префикс+суффикс | за-ST-ива | 17096 |

| # | Слово | Часть речи | Тематика | Тип заимствования | Тип словообразования | Модель словообразования | Частота |
|---|---|---|---|---|---|---|---|
| 120 | улыбизм | N | психология | Исконное | Суффикс | ST-изм | 19457 |
| 121 | коучинг | N | психология | Англицизм | | | 100986 |
| 122 | девелопер | N | работа | Англицизм | | | 22447 |
| 123 | дедлайн | N | работа | Англицизм | | | 19996 |
| 124 | корпоратив | N | работа | Англицизм | | | 39322 |
| 125 | рекрутинг | N | работа | Англицизм | | | 17694 |
| 126 | фрилансер | N | работа | Англицизм | | | 39902 |
| 127 | печенька | N | разное | Исконное | Суффикс | ST-к | 27218 |
| 128 | всечь | V | разное | Исконное | Префикс | в-ST | 24566 |
| 129 | мда | Interj | разное | Исконное | | | 51107 |
| 130 | медвед | N | разное | Исконное | | | 19568 |
| 131 | пристройство | N | разное | Исконное | Суффикс | ST-ств | 24515 |
| 132 | декупаж | N | разное | Галлицизм | | | 29524 |
| 133 | инфо | N | разное | Англицизм | | | 144808 |
| 134 | брейн | Nmod | разное | Англицизм | | | 104056 |
| 135 | вау | Interj | разное | Англицизм | | | 47633 |
| 136 | копирайтинг | N | разное | Англицизм | | | 53164 |
| 137 | лайт | Nmod/N | разное | Англицизм | | | 21867 |
| 138 | плиз | Interj | разное | Англицизм | | | 60359 |
| 139 | райдер | N | разное | Англицизм | | | 26129 |
| 140 | сексодром | N | разное | Англицизм | | | 27398 |
| 141 | сорри | Interj | разное | Англицизм | | | 27449 |
| 142 | стайл | Nmod/N | разное | Англицизм | | | 23461 |
| 143 | форсайт | Nmod | разное | Англицизм | | | 22689 |
| 144 | чарт | N | разное | Англицизм | | | 29022 |
| 145 | байк | Nmod/N | спорт и досуг | Англицизм | | | 24999 |
| 146 | билдер | N | спорт и досуг | Англицизм | | | 31338 |
| 147 | геймплей | N | спорт и досуг | Англицизм | | | 19222 |
| 148 | караванер | N | спорт и досуг | Англицизм | | | 44569 |
| 149 | клаб | Nmod | спорт и досуг | Англицизм | | | 37051 |
| 150 | холдем | N | спорт и досуг | Англицизм | | | 43273 |
| 151 | шутер | N | спорт и досуг | Англицизм | | | 28704 |
| 152 | флешка | N | техника | Смешанное | Суффикс | ST-к | 56179 |
| 153 | планшетник | N | техника | Смешанное | Суффикс | ST-ник | 25473 |
| 154 | аудиокнига | N | техника | Смешанное | Композит | ST-ST | 73058 |
| 155 | мультиварка | N | техника | Смешанное | Композит | ST-ST | 18704 |
| 156 | айпад | N | техника | Англицизм | | | 38430 |
| 157 | айфон | N | техника | Англицизм | | | 111594 |
| 158 | бумбокс | N | техника | Англицизм | | | 47473 |
| 159 | виджет | N | техника | Англицизм | | | 25507 |
| 160 | ноут | N | техника | Англицизм | | | 30447 |
| 161 | плагин | N | техника | Англицизм | | | 79222 |
| 162 | плейлист | N | техника | Англицизм | | | 256455 |
| 163 | подкаст | N | техника | Англицизм | | | 68787 |
| 164 | рингтон | N | техника | Англицизм | | | 20499 |
| 165 | роутер | N | техника | Англицизм | | | 18340 |
| 166 | креативщик | N | человек | Смешанное | Суффикс | ST-щик | 19408 |
| 167 | бухарь | N | человек | Исконное | Суффикс | ST-арь | 23749 |
| 168 | днюха | N | человек | Исконное | Суффикс | ST-юх | 18235 |